\documentclass[10pt]{article}
\usepackage[utf8]{inputenc}
\usepackage{graphicx}
\usepackage{amsfonts}
\usepackage{amsmath}
\usepackage[margin=.6in]{geometry}
\usepackage{xcolor}
\usepackage{float}

\newcommand{\norm}[1]{\left\lVert#1\right\rVert_2^2}

\title{On the role of depth predictions for 3D human pose estimation}
\author{Alec Diaz-Arias$^{1,*}$, Mitchell Messmore$^1$, Dmitriy Shin$^1$, Stephen Baek$^{1,2}$ \\  $^1$Inseer Inc. \\ $^2$University of Iowa \\ alec.diaz-arias@inseer.com}

\begin{document}

\maketitle

\begin{abstract}
    Following the successful application of deep convolutional neural networks to 2d human pose estimation, the next logical problem to solve is 3d human pose estimation from monocular images. While previous solutions have shown some success, they do not fully utilize the depth information from the 2d inputs. With the goal of addressing this depth ambiguity, we build a system that takes 2d joint locations as input along with their estimated depth value and predicts their 3d positions in camera coordinates. Given the inherent noise and inaccuracy from estimating depth maps from monocular images, we perform an extensive statistical analysis showing that given this noise there is still a statistically significant correlation between the predicted depth values and the third coordinate of camera coordinates. We further explain how the state-of-the-art results we achieve on the H3.6M validation set are due to the additional input of depth. Notably, our results are produced on neural network that accepts a low dimensional input and be integrated into a real-time system. Furthermore, our system can be combined with an off-the-shelf 2d pose detector and a depth map predictor to perform 3d pose estimation in the wild. 
\end{abstract}

\section{Introduction}
Given an image of a person, pose estimation is the process of producing a 3d skeleton that matches the spatial position and configuration of the individual. In practice, one starts with an image containing a human actor and attempts to predict the 3d coordinates of the joints from the actor's skeleton. Learning approaches fall into the dichotomy of top-down or bottom-up. Top-down approaches first create a bounding box for the actor and then apply a pose estimation module to the bounded image, while bottom-up approaches start by predicting 3d joint locations and then assigning these key-points to individual actors using clustering algorithms. Within the top-down approach there is a further bifurcation of one stage versus two stage networks. A one stage network (like \cite{MCL19}) takes as input the 2d image. On the other hand, a two stage network first extracts 2d joint information and then from this predicts the 3d data.

Depth plays an important part of the 3d reconstruction process and should implicitly be learned by a network. However, depth information has only recently \cite{PML16, M-JR-RM-SM-C18, VL20} been explicitly used as input. This successful inclusion of true depth maps has lead us to examine the effectiveness of predicted depth into 3d pose estimation.

\paragraph{Hypothesis 1: Depth is useful for two-dimensional to three-dimensional reconstruction tasks.}

Intuitively, any information that is ``orthogonal'' to the locations in pixel coordinates can naturally improve the accuracy of a reconstruction network. We hypothesize that including depth estimates for all joints will result in increased accuracy. To test this theory, we feed a modified network, based on the architecture in \cite{MHRL17}, joint locations in pixel coordinates along with estimated depth values. This network achieves state-of-the-art results on the the largest publicly available
3d pose estimation benchmark, Human 3.6M \cite{IPOS14}.

\paragraph{Hypothesis 2: Depth estimations that are more strongly correlated with the $z$-coordinates of joint locations lead to a lower avg mm error.}

A higher correlation between depth values and joint $z$-coordinates indicates a lower degree polynomial can be used to model this relationship. In this case, a lower capacity network can be used to model this dependence, which is exactly displayed in our investigations of hypothesis one. When compared to the network developed in \cite{MCL19}, our network which utilizes depth values is less complex, yet produces more accurate results.

We test and validate this hypothesis by calculating correlation and statistical significance levels for individual joints sampled by camera and action. Subsequently we compare the same joints average mm error for high correlation sub-samples versus low correlation sub-samples. This statistical analysis lends credence to hypothesis. We discuss short comings of the data such as occlusions, and suggest directions to further the explore this theory. Our experiments lead us to the belief that the community should utilize correlation as a metric by which to gauge the efficacy of depth estimators as to not lose sight of depth maps desired utility. 

The results of our investigation emphasize the role of depth in 3d reconstruction. Our contributions can be summarized as follows:
\begin{enumerate}
    \item We introduce orthogonal data input (depth values) to supplement the 2d data inputs into our 3d human pose estimation model.
    \item We achieve state of the art 3d pose estimation results on both protocols for H3.6M (with and without alignment), using a network that can be run in real-time in conjunction with an off the shelf 2d pose estimator \cite{CH-MSWS19,WRKS16}.
    \item We conduct an extensive correlation analysis that motivated and justifies the use of depth information extracted from an RGB image.
    \item We show that our network significantly outperforms previous methods, achieving up to a 38 percent MPJPE reduction on protocol 2 with no rigid alignment on the Human 3.6M data.
\end{enumerate}

\section{Previous Work}
 The introduction of deep convolutional neural networks has led to a steady increase in performance of depth map estimation, but significant problems still exist both in quality and resolution of the depth maps. We do not seek to improve upon the state of the art in the area of depth estimation, but we aim to demonstrate that substantial improvement in this area would lead to large impact in the area of 3d human pose estimation.

3d pose estimation is divided into two major camps, single and two staged approaches. The single stage attempts to reconstruct the three-dimensional skeleton directly from the input image; while the two stage methods utilize the high accuracy of 2d pose estimators to first locate the skeleton in pixel coordinates and then learn the camera intrinsic parameters as well as depth, indirectly. \cite{LXSCLH14,PZDD17,MCL19} are based on the single stage approach while \cite{MHRL17,CL17,CR17,FXWLZ18,PHK16,YOWRLW18} are based on the two stage approach. 

\cite{LXSCLH14} proposed a multitask framework that simultaneously trains the reconstruction and the pixel coordinate detection network. \cite{TKSLF16} introduced an over-complete autoencoder to embed the skeletal representation in a high dimensional space, and this work has influenced our choice to embed in a high dimensional space. \cite{SSLW17} introduced what he called a compositional loss to encourage the network to pay attention to the joint connection structure. \cite{PHK16,MHRL17} both estimate the 2d pose from the image and then directly regress the 3d pose. \cite{MHRL17} proposed a simple architecture that was state-of-the-art and has significantly influenced our design choice. \cite{MCL19} introduces a novel approach to multi-person pose estimation by performing reconstruction on each subject individually in the root centered camera reference frame while simultaneously learning the absolute root positions to perform translation. 

There is a distinct approach to all of the above. In the above papers and as well as in our work, the networks directly estimate the joint locations in Cartesian coordinates. It is well known that points in three-space admit many representations, e.g. Euler angles. 
\cite{BK01,BKLGRB16,PC04,ZSZLW16} choose to estimate the angular representation of the skeleton directly from RGB images. This on one hand has a possible advantage; not all joints have 3 degrees of freedom in the angular representation -- thus, they have lower dimensionality. Furthermore, these approaches are less susceptible to varying limb lengths. We have not experimented with this representation of data, as our depth estimates that increase the accuracy of the network are related to the $z$-coordinate in the camera reference frame. 

Lastly, there has been recent work \cite{PML16, M-JR-RM-SM-C18, VL20} in this direction utilizing depth maps to aid in three-dimensional human pose estimation although their approaches have differed any key ways. \cite{PML16} utilizes only a single view depth map as an input and instead of directly regressing the 3d joint locations applies an iterative update to some mean pose. \cite{VL20} introduces a notion of weak depth supervision, i.e. a model that can accept either RGB or RGB-D images as inputs and achieved SOTA on MuPots by using a robust occlusion loss. \cite{M-JR-RM-SM-C18} introduces a deep neural network called Deep Depth Pose and directly regresses the camera centered three-dimensional pose. In our view, there are a couple main distinctions between our approach and our predecessors. First, none of these approaches are utilizing predicted depth maps and thus do not have the in-depth statistical analysis. In our view this reduces the problem to 2d human pose estimation alongside learning the camera instrinsics, rather than focusing on the interpolation between predicted depth and the true depth. Second, the 3d pose estimation networks themselves in these examples are not light-weight in the sense that they use RGB-D or D images as their input thus increasing the dimensionality. %However, on the other hand they do not suffer from needing multiple networks interacting simultaneously. 

\section{Methodology}
Our goal is to estimate three-dimensional body joint locations given a three-dimensional input where the third dimension of the input is ``orthogonal'' to the pixel coordinate joint locations. We will argue later that this increased dimensionality is critical to our network's success. Additionally, we show this extra dimension is correlated with the $z$-coordinate, thus lending credence to its utility. Formally, let $x \in \mathbb{R}^{3J}$ denote our input and denote the output as $y \in \mathbb{R}^{3J}$, where $J$ is the number of joints to be estimated. We seek to learn a function $f: \mathbb{R}^{3J} \to \mathbb{R}^{3J}$ that minimizes the joint reconstruction error:
\[
\min_{f}{\frac{1}{J}\sum\limits_{i=1}^{J}{||f(x_i)-y_i||_2^2}} \text{.}
\]

In practice, $x_i$ may be obtained using an off-the-shelf 2d pose detector \cite{CH-MSWS19,WRKS16} and a depth map estimation algorithm using monocular images, i.e. let $(u,v)$ denote the pixel coordinate location of joint $j_{i}$ and let $D$ denote the depth map, then $x_i = (u,v,D(u,v))$.  We estimate the 3d joint locations in the camera reference frame. We aim to approximate the reconstruction function $f$ using a neural networks as a function estimator.

\begin{figure}[H]

  \includegraphics[width=\textwidth]{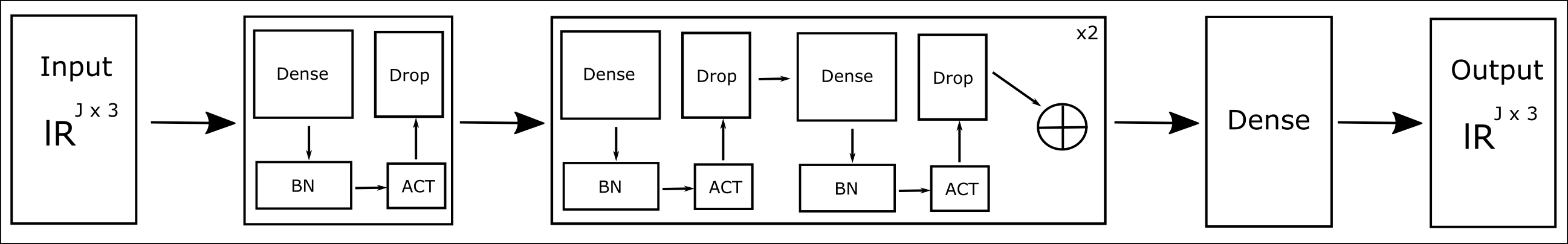}
  \label{fig:architecture}
  \caption{The proposed 3d pose estimation network architecture. Where BN denotes Batch-Normalization and Act denotes non-linear activation function}
\end{figure}

Figure~\ref{fig:architecture} provides a diagram with the basic building blocks of our network. Our network is modeled after \cite{MHRL17} e.g. using low dimensional input (i.e. the Cartesian coordinates of the skeleton) and is based on their deep multi-layer neural network with batch-normalization (BN) and dropout (Drop) to reduce over fitting. Our network has approximately 7 million parameters, but due to the low dimensional input the network is easily integrated into a real-time system. 

\subsection{Pixel, Camera, and World Coordinates}
Recall that given a point  $P_{u,v,w}$ in the world reference frame (homogeneous coordinates), the camera intrinsic/extrinsic parameters $R \in SO(3)$ and $t \in \mathbb{R}^3$,  and a projection $P \in M_{3}(\mathbb{R})$,  one can move to the camera reference frame by $P_{x,y,z}=RP_{u,v,w} +t$. Furthermore, moving to pixel coordinates amounts to applying the following transformation (the perspective projection): 
$P_{r,s}=P_{per}(P_{x,y,z})$ where 
\[
P_{per}=\begin{bmatrix}
f_x & 0 & c_x \\
0 & f_y & c_y \\
0 & 0 & 1
\end{bmatrix} \text{,}
\]
where $f_x=\frac{f}{s_x}$ and $f_y=\frac{f}{s_y}$ and $f$ is the focal length, $s_x$ and $s_y$ are the effective size, and $c_x$ and $c_y$ describes the principle point.
Since the pre-image of $P_{per}^{-1}((u,v))$ is uncountable for any pixel coordinate our problem is extremely ill-posed, there is no analytical solution. We do note that converting back from pixel-coordinates to the camera reference frame can be achieved using the following system of equations and are derived from the above equations:

\begin{gather*}
 x= \frac{f_x}{z}(r-c_x)\\
 y= \frac{f_y}{z}(s-c_y) \\
 z = z_{depth}.
\end{gather*}
The above clearly emphasizes the above statement that our problem is ill-posed. First, there are four intrinsic parameters required for the reconstruction (assuming only one camera is used) but the z-value varies depending on the pixel coordinate inputs. This is the major hurdle, in our opinion, to achieve high quality reconstruction. We believe independent of the complexity or novelty of the neural network topology these networks will under-perform any counterpart using additional input that is ``orthogonal'' to the pixel level information. In the presence of the true depth value the problem is still ``ill-posed'' but significantly more tractable. Furthermore, what we exploit is a noisy depth value that at most has varying correlation for differing joints when considered across our sampling procedure (moderately high to weak). Nonetheless we still achieve state-of-the-art results on the H36M dataset.

\subsection{Depth Estimation}
We use a simple encoder-decoder architecture to estimate depth, which is a common practice in depth estimation \cite{GKCR16}. We train on the publicly available NYU V2 \cite{SHKF12} dataset where the ground truth depth maps are generated using the Kinect V2. Recall that the focus of this paper was not to improve upon the current state of the art, rather to generate depth maps with some correlation and make a strong case for this being a missing component to the reconstruction problem.  We simply note that even with sub optimal depth maps/values a simple neural network with the additional orthogonal input can outperform the previous state-of-the-art in the 3d human pose estimation. Therefore, 3d pose estimation can be further improved by creating better quality depth maps, \textit{i.e.} depth map with stronger correlations with the $z$-coordinate from the camera coordinate system. 

\begin{figure}[H]
\centering
\caption{Example of depth map predicted on the H36M dataset}
\includegraphics[scale=.2]{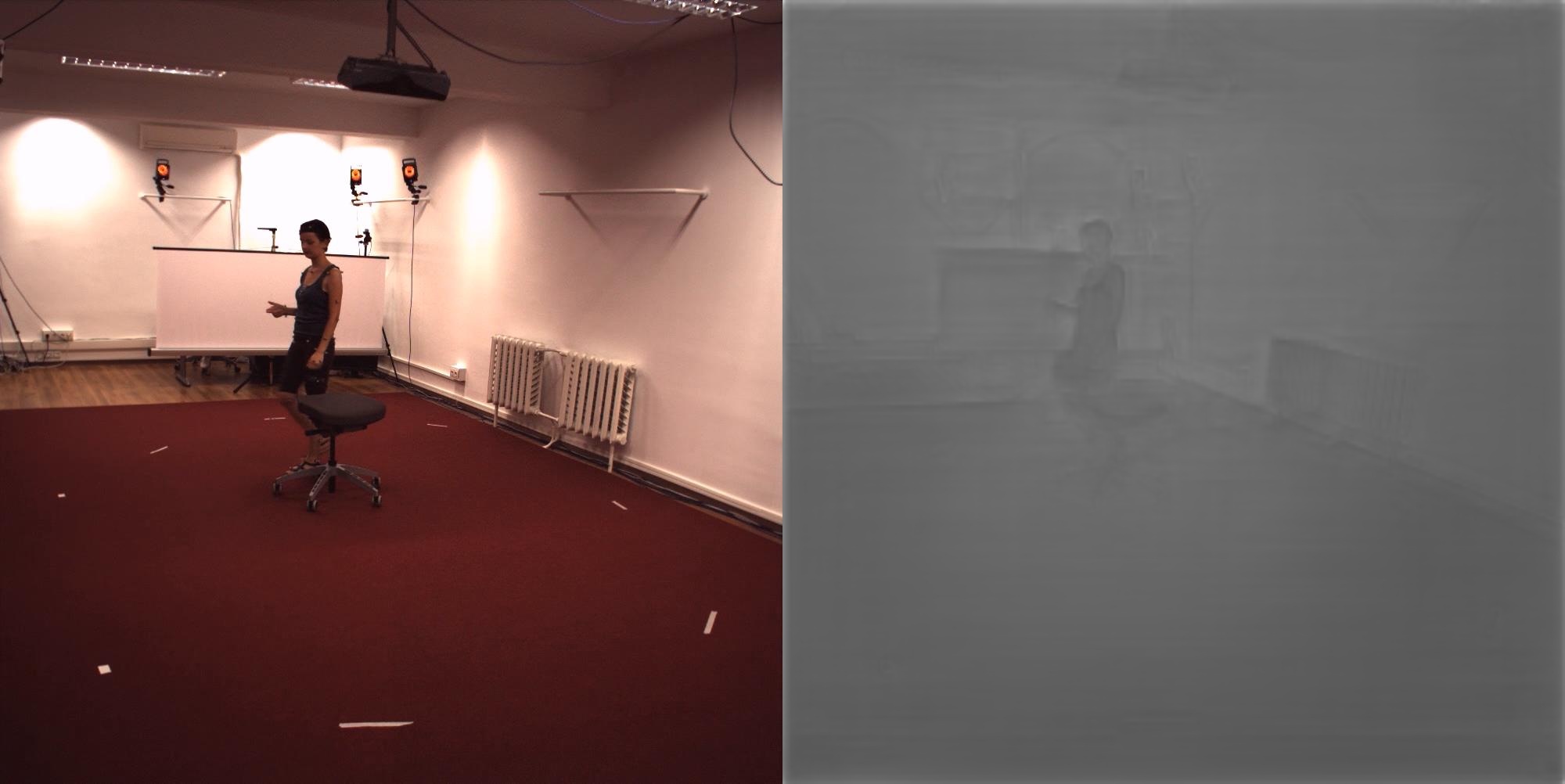}
\end{figure}

We train the depth estimation network with respect to the following losses:
\[
L_\text{MSE}(y,\hat{y})=\frac{1}{n}\sum\limits_{p=1}^{n}{\norm{y-\hat{y}}}
\]
and the gradient loss defined over the depth image, \textit{i.e.},
\[
L_\text{grad}(y,\hat{y})=\frac{1}{n}\sum\limits_{p=1}^{n}{\norm{\nabla_x(y_p,\hat{y}_p)} + \norm{\nabla_y(y_p,\hat{y}_p)} } \text{,}
\]
where $y$ is the ground truth, $\hat{y}$ is the predicted output and $\nabla_{x}$ and $\nabla_{y}$ are the gradients with respect to $x$ and $y$, respectively.

\subsection{Data preprocessing}
We apply normal standardization to the 2d inputs and 3d outputs of the network. We subtract the mean pose and divide by its standard deviation pose of the training dataset. We also do the same normalization on the depth values. We also zero center the root joint (pelvis).

\subsection{Training Details}
We train our network for only 70 epochs using an Adam optimizer with an initial learning rate of 0.01, and a batch size of 1,024. Initially, the weights are generated using the Xavier uniform initialization scheme \cite{GB10}. We implemented our code using native TensorFlow \cite{TF15} which takes around 392ms for a forward and backward pass per batch, and was trained using a NVIDIA GeForce RTX 2080 Max Q design. The forward pass takes around 4 ms on the same RTX 2080 for a single sample.  This implies that our network used in conjunction with a standard off the shelf 2d pose detector could be implemented in real time as a pixels to 3d coordinates system. One epoch of training on the entire Human 3.6M dataset takes roughly 10 minutes. Hence our network achieves state of the art results while being a lower capacity network.

\section{Quantitative results}
Before moving to the numerical results, we first provide examples of our reconstruction relative to the ground truth from differing camera angles and distinct poses. 
\begin{figure}[H]
  \caption{Blue skeleton is the ground truth pose from H36M while the red is the predicted skeleton with our network.}
  \includegraphics[width=\textwidth]{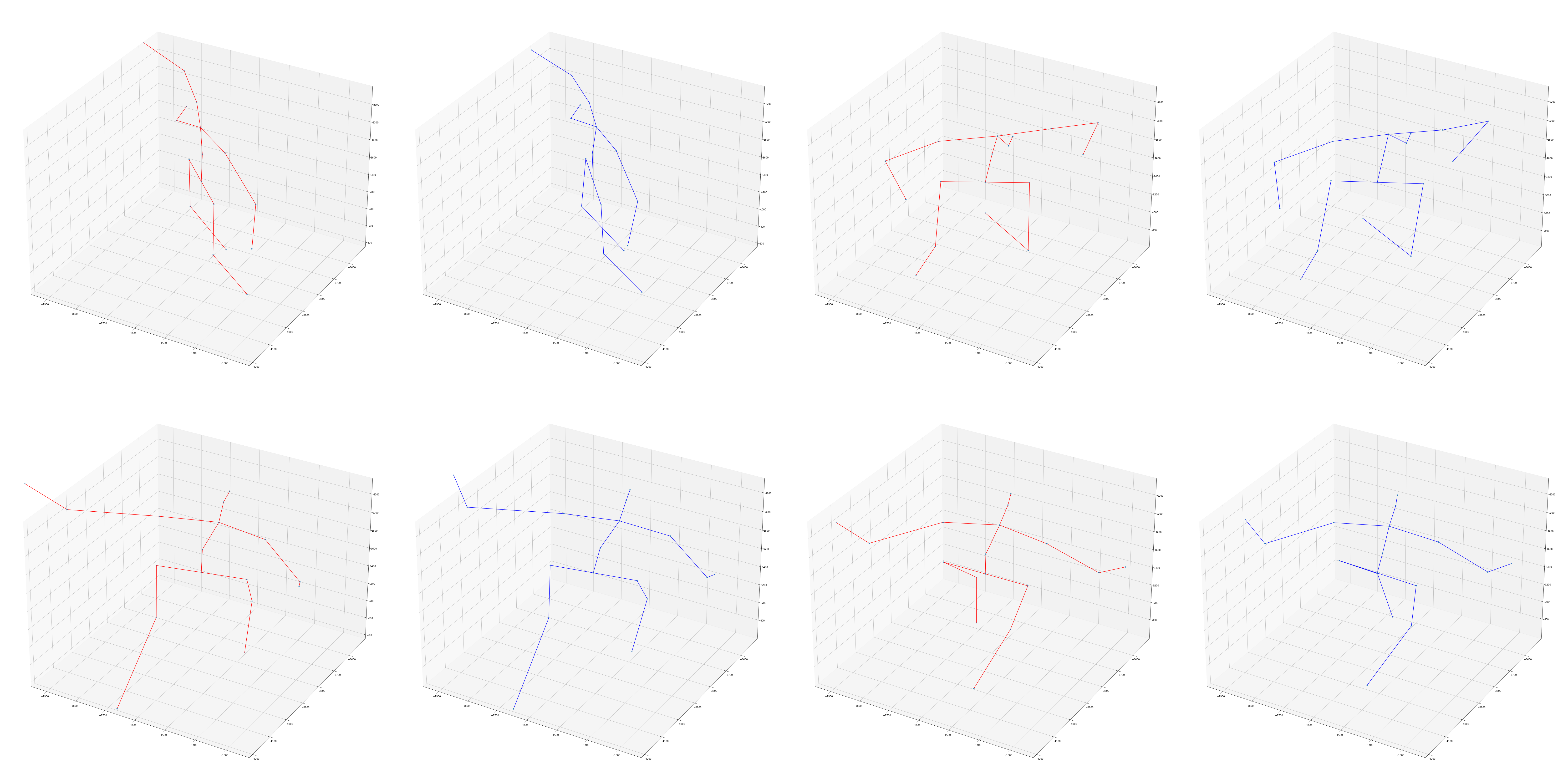}
\end{figure}

\subsection{Experimental protocol}
\textbf{Human 3.6M dataset.} Two experimental protocols are widely used on the H3.6M dataset \cite{MCL19,FXWLZ18,MHRL17,ZHSXW17,M17,CR17,YIKWG16,SSLW17,SXWLW18}. Protocol 1 uses six subjects for training (S1,S5,S6,S7,S8,S9) and one for testing (S11). The metric we use is mean per joint position error (MPJPE) \cite{{ZC15}} for evaluation. Protocol 2 uses five subjects for training (S1,S5,S6,S7,S8) and two for testing (S9,S11) along with the same evaluation metric.  We use every 64th frame in the videos for testing following \cite{SSLW17,SXWLW18}. Finally, we report the MPJPE in two cases, the first is without ground truth information during testing, \textit{i.e.} rigid alignment and the other is with rigid alignment after prediction. 

We perform rigid alignment by finding $R \in SO(3)$ and $t \in \mathbb{R}^3$ such that $\sum\limits_{i=1}^{J}\norm{(Rp_i+t)-y_i}$ is minimized where $p_i$ is the predicted joint location and $y_i$ is the ground truth. Tables \ref{Table:QuantRes1} and \ref{Table:QuantRes2} below provide a comparison between previous methods' errors on the H3.6M data subdivided by each action and then the average error is provided. Lastly, we subdivide the results when ground truth information is provided (W GT \textit{i.e.} rigid alignment) and without (W/O GT).

\begin{table}[H] 
\centering
 \begin{tabular}{ | p{.7cm} p{.7cm} p{.7cm} p{.7cm} p{.7cm} p{.7cm} p{.7cm} p{.7cm} p{.7cm} p{.7cm} p{.7cm} p{.7cm} p{.7cm} p{.7cm} p{.9cm} p{.7cm} p{.7cm} | } 
 \hline
 Joint & RH & RK & RA & LH & LK & LA & Tho. & Neck & Nose & Head & LS & LE & LW & RS & RE & RW \\
 \hline

 P1 & 9.99 & 29.51 & 48.79 & 9.99 & 32.53 & 52.91 & 20.84 & 27.54 & 33.85 & 39.14 & 28.29 & 47.32 & 68.47 & 29.64 & 50.21 & 63.99  \\
 
  P2 & 14.79 & 35.14 & 55.88 & 14.79 & 38.77 & 64.03 & 23.52 & 29.58 & 37.22 & 41.32 & 37.07 & 49.53 & 68.92 & 38.58 & 55.86 & 70.82 \\
 \hline
  \end{tabular}
  
  \caption{Average millimeter error per joint for protocol 1 (P1) and protocol 2 (P2) on H3.6M data \cite{IPOS14}. RH, RK, RA, LH, LK, and LA stand for (right-left) hip, knee, and ankle respectively. RS, RE, RW, LS, LE and LW stand for (right-left) shoulder, elbow and wrist respectively.}
  \label{Table:QuantRes3}
\end{table}

\begin{table}[H] 
\centering
 \begin{tabular}{ | p{1.9cm} p{.5cm} p{.5cm} p{.5cm} p{.5cm} p{.75cm} p{.5cm} p{.5cm} p{.5cm} p{.75cm} p{.5cm} p{.75cm} p{.75cm} p{.75cm} p{.9cm} p{.9cm} p{.75cm} | } 
 \hline
Methods & Dir. & Dis. & Eat & Gre. & Phon. & Pose & Pur. & Sit. & SitD. & Smo. & Phot. & Wait & Walk & WalkD. & WalkP. & Avg \\
 \hline\hline
 W GT  & & & & & & & & & & & & & & & & \\
 \hline
     Yasin \cite{YIKWG16} & 88.4 & 72.5 & 108.5 & 110.2 & 92.1 & 81.6 & 107.2 & 119 & 170.8 & 108.2 & 142.5 & 86.9 & 92.1 & 165.7 & 102.0 & 108.3 \\
     Chen \cite{CR17} & 71.6 & 66.6 & 74.7 & 79.1 & 70.1 & 67.6 & 89.3 & 90.7 & 195.6 & 83.5 & 93.3 & 71.2 & 55.7 & 85.9 & 62.5 & 82.7 \\
     Moreno \cite{M17} & 67.4 & 63.8 & 87.2 & 73.9 & 71.5 & 69.9 & 65.1 & 71.7 & 98.6 & 81.3 & 93.3 & 74.6 & 76.5 & 77.7 & 74.6 & 76.5 \\
     Zhou \cite{ZHSXW17} & 47.9 & 48.8 & 52.7 & 55.0 & 56.8 & 49 & 45.5 & 60.8 & 81.1 & 53.7 & 65.5 & 51.6 & 50.4 & 54.8 & 55.9 & 55.3  \\
     Mar. \cite{MHRL17} & 39.5 & 43.2 & 46.4 & 47 & 51.0 & 41.4 & 40.6 & 56.5 & 69.4 & 49.2 & 56.0 & 45.0 & 38.0 & 49.5 & 43.1 & 47.7  \\
     Sun \cite{SSLW17} & 42.1 & 44.3 & 45.0 & 45.4 & 51.5 & 43.2 & 41.3 & 59.3 & 73.3 & 51.0 & 53.0 & 44.0 & 38.3 & 48.0 & 44.8 & 48.3 \\
     Fang \cite{FXWLZ18} & 38.2 & 41.7 & 43.7 & 44.9 & 48.5 & 40.2 & 38.2 & 54.5 & 64.4 & 47.2 & 55.3 & 44.3 & 36.7 & 47.3 & 41.7 & 45.7 \\
     Sun \cite{SXWLW18} & 36.9 & 36.2 & 40.6 & 40.4 & 41.9 & 34.9 & 35.7 & 50.1 & 59.4 & 40.4 & 44.9 & 39.0 & 30.8 & 39.8 & 36.7 & 40.6  \\
     Moon \cite{MCL19} & 31.0 & 30.6 & 39.9 & 35.5 & 34.8 & 30.2 & 32.1 & 35.0 & 43.8 & 35.7 & \textbf{37.6} & 30.1 & 24.6 & \textbf{35.7} & 29.3 & 34.0 \\ 
     Ours & \textbf{23.5} & \textbf{27.2} & \textbf{27.6} & \textbf{27.2} & \textbf{26.2} & \textbf{27.9} & \textbf{24.8} & \textbf{27.9} & \textbf{41.4} & \textbf{32.9} & 39.4 & \textbf{28.6} & \textbf{20.1} & 37.2 & \textbf{25.0} & \textbf{29.1} \\
     
     \hline
     W/O GT & & & & & & & & & & & & & & & &  \\
     \hline
     Rogez \cite{RWS2019} & & & & & & & & & & & & & & & & 42.7 \\
     Moon \cite{MCL19} & 32.5 & \textbf{31.5} & 41.5 & 36.7 & 36.3 & \textbf{31.9} & 33.2 & 36.5 & \textbf{44.4} & 36.7 & \textbf{38.7} & \textbf{31.2} & 25.6 & \textbf{37.1} & 30.5 & 35.2 \\
     Ours & \textbf{29.9} & 33.1 & \textbf{32.3} & \textbf{30.8} & \textbf{33.5} & 37.1 & \textbf{27.1} & \textbf{34.5} & 48.7 & \textbf{35.7} & 49.3 & 37.3 & \textbf{23.5} & 40.6 & \textbf{27.8} & \textbf{34.8} \\
 \hline
 \end{tabular}
 
 \caption{Detailed results on the H3.6M dataset \cite{IPOS14} under protocol 1. We report with rigid alignment (W GT) and without rigid alignment (W/O GT). MPJPE is subdivided by action and the average across all actions is provided.  We underline the values that are the best. }
\label{Table:QuantRes1}
\end{table}

\begin{table}[H]
\centering
 \begin{tabular}{ | p{1.9cm} p{.5cm} p{.5cm} p{.5cm} p{.5cm} p{.75cm} p{.5cm} p{.5cm} p{.5cm} p{.75cm} p{.5cm} p{.75cm} p{.75cm} p{.75cm} p{.9cm} p{.9cm} p{.75cm} | } 
 \hline
Methods & Dir. & Dis. & Eat & Gre. & Phon. & Pose & Pur. & Sit. & SitD. & Smo. & Phot. & Wait & Walk & WalkD. & WalkP. & Avg \\
 \hline\hline
    W GT & & & & & & & & & & & & & & & & \\
    
    \hline
    
     Chen \cite{CR17} & 89.9 & 97.6 & 90.0 & 107.9 & 107.3 & 93.6 & 136.1 & 133.1 & 240.1 & 106.7 & 139.2 & 106.2 & 87.0 & 114.1 & 90.6 & 114.2 \\
     Tome \cite{TRA17} & 65.0 & 73.5 & 76.8 & 86.4 & 86.3 & 68.9 & 74.8 & 110.2 & 173.9 & 85.0 & 110.7 & 85.8 & 71.4 & 86.3 & 73.1 & 88.4 \\
     Moreno \cite{M17} & 69.5 & 80.2 & 78.2 & 87.0 & 100.8 & 76.0 & 69.7 & 104.7 & 113.9 & 89.7 & 102.7 & 98.5 & 79.2 & 82.4 & 77.2 & 87.3 \\
     Zhou \cite{ZHSXW17} & 68.7 & 74.8 & 67.8 & 76.4 & 76.3 & 84.0 & 70.2 & 88.0 & 113.8 & 78.0 & 98.4 & 90.1 & 62.6 & 75.1 & 73.6 & 79.9 \\
     Jah. \cite{JY17} &  74.4 & 66.7 & 67.9 & 75.2 & 77.3 & 70.6 & 64.5 & 95.6 & 127.3 & 79.6 & 79.1 & 73.4 & 67.4 & 71.8 & 72.8 & 77.6 \\
     Mehta \cite{MSMXSMT18} & 57.5 & 68.6 & 59.6 & 67.3 & 78.1 & 56.9 & 69.1 & 98.0 & 117.5 & 69.5 & 82.4 & 68.0 & 55.3 & 76.5 & 61.4 & 72.9 \\
     Mar. \cite{MHRL17} & 51.8 & 56.2 & 58.1 & 59.0 & 69.5 & 55.2 & 58.1 & 74.0 & 94.6 & 62.3 & 78.4 & 59.1 & 49.5 & 65.1 & 52.4 & 62.9  \\
     Fang \cite{FXWLZ18} & 50.1 & 54.3 & 57.0 & 57.1 & 66.6 & 53.4 & 55.7 & 72.8 & 88.6 & 60.3 & 73.3 & 57.7 & 47.5 & 62.7 & 50.6 & 60.4 \\
     
     Sun \cite{SSLW17} & 52.8 & 54.8 & 54.2 & 54.3 & 61.8 & 53.1 & 53.6 & 71.7 & 86.7 & 61.5 & 67.2 & 53.4 & 47.1 & 61.6 & 63.4 & 59.1 \\

     Sun \cite{SXWLW18} & 47.5 & 47.7 & 49.5 & 50.2 & 51.4 & 43.8 & 46.4 & 58.9 & 65.7 & 49.4 & 55.8 & 47.8 & 38.9 & 49.0 & 43.8 & 49.6  \\
     Moon \cite{MCL19} & 50.5 & 55.7 & 50.1 & 51.7 & 53.9 & 46.8 & 50.0 & 61.9 & 68.0 & 52.5 & 55.9 & 49.9 & 41.8 & 56.1 & 46.9 & 53.3 \\
     Ours & \textbf{25.2} & \textbf{29.0} & \textbf{30.8} & \textbf{28.9} & \textbf{28.6} & \textbf{29.8} & \textbf{26.1} & \textbf{30.3} &  \textbf{43.7} & \textbf{34.1} & \textbf{41.3} & \textbf{30.4} & \textbf{21.6} & \textbf{36.3} & \textbf{26.9} & \textbf{33.3} \\

    \hline
     W/O GT & & & & & & & & & & & & & & & &  \\
     \hline
     Rogez \cite{RWS17} & 76.2 & 80.2 & 75.8 & 83.3 & 92.2 & 79.9 & 71.7 & 105.9 & 127.1 & 88.0 & 105.7 & 83.7 & 64.9 & 86.6 & 84.0 & 87.7 \\
     Mehta \cite{MSMXSMT18} & 58.2 & 67.3 & 61.2 & 65.7 & 75.8 & 62.2 & 64.6 & 82.0 & 93.0 & 68.8 & 84.5 & 65.1 & 57.6 & 72.0 & 63.6 & 69.9 \\
     Rogez \cite{RWS2019} & 55.9 & 60.0 & 64.5 & 56.3 & 67.4 & 71.8 & 55.1 & 55.3 & 84.8 & 90.7 & 67.9 & 57.5 & 47.8 & 63.3 & 54.6 & 63.5 \\
     Moon \cite{MCL19} & 51.5 & 56.8 & 51.2 & 52.2 & 55.2 & 47.7 & 50.9 & 63.3 & 69.9 & 54.2 & 57.4 & 50.4 & 42.5 & 57.5 & 47.7 & 54.4   \\
     Ours & \textbf{32.9}  & \textbf{35.9} & \textbf{35.1} & \textbf{34.5} & \textbf{37.8} & \textbf{41.0} & \textbf{30.0} & \textbf{38.2} & \textbf{51.2} & \textbf{37.7} & \textbf{52.2} & \textbf{39.2} & \textbf{25.8} & \textbf{41.4} & \textbf{30.6} & \textbf{39.7} \\
 \hline
 \end{tabular} 
 
 \caption{Detailed results on H3.6M dataset \cite{IPOS14} protocol 2. We report with rigid alignment (W GT) and without rigid alignment (W/O GT). MPJPE is subdivided by action and the average across all actions is provided. We put in bold the best results.}
\label{Table:QuantRes2}
\end{table}

Next, we move to the correlation analysis that led us to using depth values of specific joints as additional input. While noisy depth has intuitive utility and ground truth depth maps have been used in the literature -- it was unclear to what extent noisy depth maps could aid in the reconstruction problem.

\section{Correlation Analysis of Predicted Depth vs $Z$-coordinate}
We performed a statistical analysis to assess the extent of possible correlation between depth values and the $z$-coordinate of joints in the camera reference frame. We choose to sub-sample our data by camera and action as this allows us to observe the extremes of correlation values. This method also better indicates the affect of joint occlusion in the reconstruction. Thus, we believe to get an accurate picture of how much correlation is present this is a natural sampling procedure. 

First, we explored descriptive statistics of the distributions to check for the normality assumption, in the case that the distribution is non-normal then we are forced to apply non-parametric methods. Since accuracy of different normality tests can depend on sample size and other characteristics of the distribution, we decided to perform several tests, namely the Shapiro-Wilk, Andersen-Darling and D’Agostino tests \cite{SW65,D52,D70} for normality. The tests were done for data samples of sizes 500, 1,000, 5,000 and 100,000. 

Tables \ref{Table:Stats1} and \ref{Table:Stats2} present the results of the tests for a distribution of 5,000 depth values by specific joints in an attempt to achieve most accurate $p$-values. As shown in the Shapiro-Wilk and D’Agostino, the reported p-values warranted rejection of the null hypothesis of the normality assumption, \textit{i.e.}, the p-values of almost all tests were lower than confidence level $\alpha=0.05$. There were only two cases when D’Agostino test produced p-values that indicated that the distribution was normal (see underlined values in Table \ref{Table:Stats1} and \ref{Table:Stats2}). All values of statistic in Andersen-Darling tests were greater than corresponding critical values, which indicated violation of normality assumption. The consensus conclusion was that the distribution of depth values was not normally distributed. This conclusion was also supported by skewness and kurtosis. All kurtosis tests confidently rejected the null hypothesis that the shape of the distribution matched the shape of the normal distribution (\textit{e.g.} peaked shape with light tails). It can also be seen from histograms in Figure 5 that some distributions were multi-modal.

\begin{table}[H]
\centering
 \begin{tabular}{ | p{1.5cm} p{.9cm} p{.9cm} p{.9cm} p{.9cm} p{.9cm} p{.9cm} p{.9cm} p{.9cm} | } 
 
  \hline
Joint & Root & RH & RK & RA & LH & LK & LA & Thor. \\

 \hline\hline
  Shapiro & 0.99 & 0.97 & 0.99 & 0.98 & 0.99 & 0.98 & 0.98 & 0.98 \\
 p-value & 3e-6 & 0.0 & 1e-4 & 0.0 & 3e-6 & 0.0 & 0.0 & 0.0 \\
 Andersen & 3.09 & 7.58 & 1.99 & 5.69 & 3.31 & 4.93 & 4.46 & 4.78 \\
 p-value & .78 & .78 & .78 & .78 & .78 &.78 & .78 & .78 \\
 D'Agostino & 12.64 & 68.5 & 1.94 & 86.7 & 10.65 & 23.9 & 53.8 & 13.4 \\
 p-value & 2e-3 & 0.0 & \underline{.37} & 0.0 & 5e-3 & 6e-6 & 0.0 & 1e-3 \\
 
  \hline
 \end{tabular} \caption{Result of statistical tests for normality for depth values by joints. Upper value for all tests is test statistic. Lower value for Shapiro-Wilk and D'Agostino tests is $p$-value and for Andersen-Darling test is critical value.}\label{Table:Stats1}
\end{table}

\begin{table}[H]
\centering
 \begin{tabular}{ | p{1.5cm} p{.9cm} p{.9cm} p{.9cm} p{.9cm} p{.9cm} p{.9cm} p{.9cm} p{.9cm} p{.9cm} | } 
 
  \hline
Joint & Neck & Nose & Head & LS & LE & LW & RS & RE & RW \\

 \hline\hline
Shapiro & 0.98 & 0.97 & 0.94 & 0.97 & 0.99 & 0.98 & 0.96 & 0.92 & 0.91 \\
p-value & 2e-6 & 0.0 & 0.0 & 0.0 & 4e-5 & 0.0 & 0.0 & 0.0 & 0.0 \\
Andersen & 4.17 & 10.4 & 24.3 & 9.9 & 2.67 & 4.18 & 12.94 & 29.92 & 29.04 \\
p-value & .78 & .78 & .78 & .78 & .78 & .78 & .78 & .78  & .78\\
D'Agostino & 9.87 & 44.88 & 64.91 & 31.61 & 4.19 & 63.12 & 39.03 & 88.54 & 89.22 \\
p-value & .007 & 0.0 & 0.0 & 0.0 & \underline{.12} & 0.0 & 0.0 & 0.0 & 0.0 \\
 
  \hline
 \end{tabular} \caption{Result of statistical tests for normality for depth values by joints cont.}\label{Table:Stats2}
\end{table}

Normality tests for a distribution of $z$-coordinate values produced similar results, and we concluded that the distribution was not normal. As a result of these tests, we were forced to apply non-parametric methods to assess the extent of correlation between depth values and $z$-coordinates. For this, we used Spearman’s rank correlation and Kendall’s Tau rank correlation tests.

\begin{figure}[H]
\caption{Histograms of depth values to assess normality}
\centering
\includegraphics[scale=0.5]{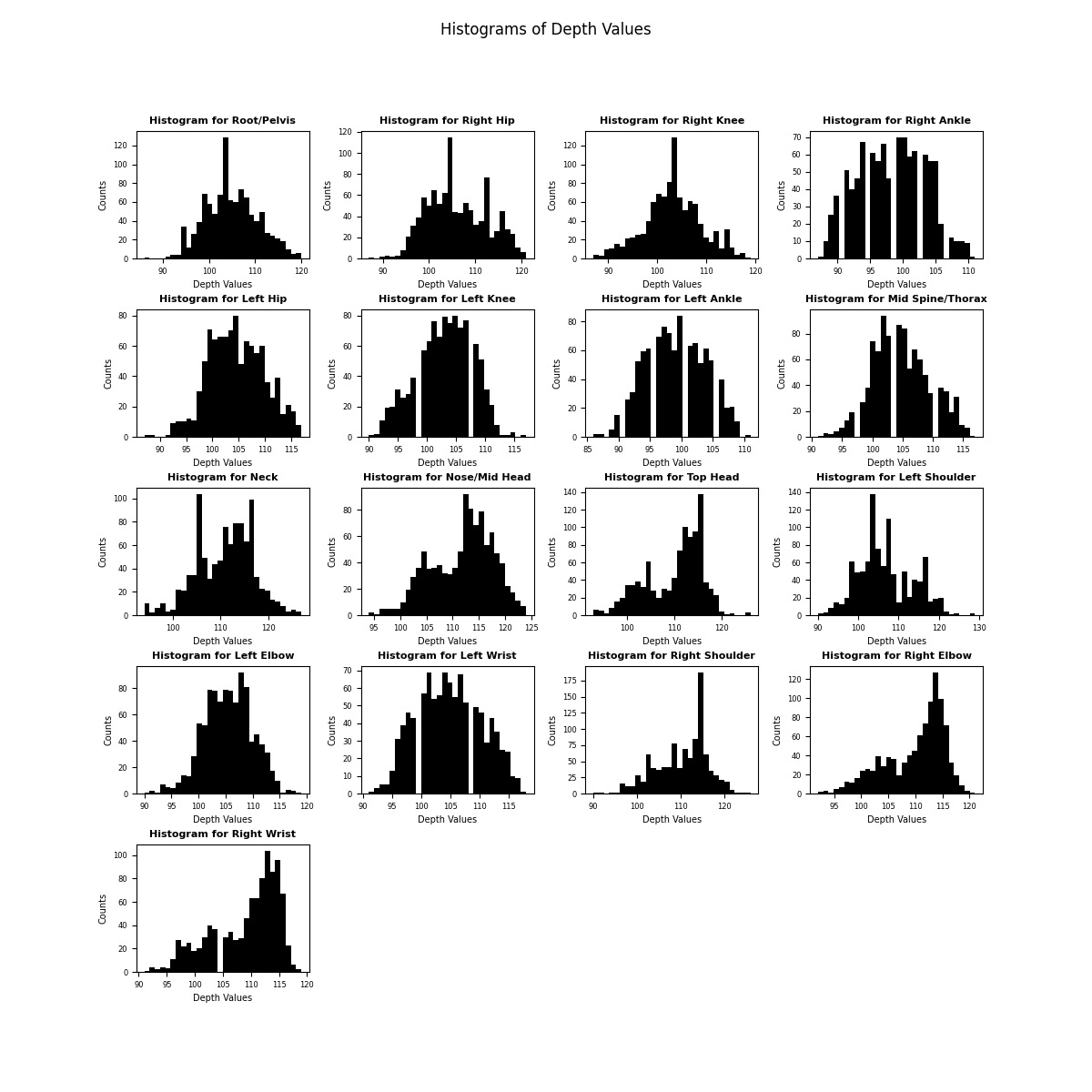}
\end{figure}

\begin{table}[H] 
\centering
 \begin{tabular}{ | p{1.5cm} p{1.5cm} p{1.5cm} p{1.5cm} p{1.5cm} p{1.5cm} p{1.5cm}  | } 

  \hline
Jt. (C,A) & RK(1,14) & RK(4,16) & LK(3,14) & LA(4,16) & LK(3,12) & RK(3,14)  \\

 \hline\hline
  Spear. & .5925 & .5692 & .5574 & .5495 & .5487 & .5311   \\
 p-value & 0.0 & 0.0 & 0.0 & 0.0 & 0.0 & 0.0  \\
 Kendall & .4086 & .3838 & .3903 & .3817 & .3955 & .3791 \\
 p-value & 0.0 & 0.0 & 0.0 & 0.0 & 0.0 & 0.0  \\
 
  \hline
 \end{tabular} \caption{Examples of moderately high Spearman and Kendall Tau rank correlation. RK, LK, and LA denote right-knee, left-knee, and left-ankle respectively and (C,A) denote camera and action.}
 \label{Table:Stats3}
\end{table}

\begin{table}[H] 
\centering
 \begin{tabular}{ | p{1.5cm} p{1.5cm} p{1.5cm} p{1.5cm} p{1.5cm} p{1.5cm} p{1.5cm}  | } 
  \hline
Jt. (C,A) & RA(2,5) & LA(1,10) & RW(3,3) & LW(1,5) & LE(2,6) & LE(2,4) \\

 \hline\hline
Spear. & .0102 & .0097 & .0095 & .0089 & .0085 & .0078 \\
p-value & .51 & .52 & .32 & .56 & .47 & .57  \\
Kendall &  .0080 & .0115 & .0058 & .0088 & .0081 & .0081\\
p-value & .45 & .26 & .36 & .41 & .32 & .38 \\
  \hline
 \end{tabular} \caption{Examples of weak Spearman and Kendall Tau rank correlation correlation that are statistically insignificant. RA, LA, RW, LW, and LE denote right-ankle, left-ankle, right-wrist, left-wrist, and left-elbow respectively and (C,A) denotes the camera and action. }
 \label{Table:Stats4}
\end{table}

\begin{table}[H] 
\centering
 \begin{tabular}{ | p{1.5cm} p{1.5cm} p{1.5cm} p{1.5cm} p{1.5cm} p{1.5cm} p{1.5cm}  | } 
  \hline
Jt. (C,A) & TH(4,2) & RH(2,10) & LE(3,9) & RK(1,9) & MH(4,2) & RW(4,2) \\

 \hline\hline
Spear. & -.526 & -.489 & -.462 & -.459 & -.457 & -.455 \\
p-value & 0.0 & 0.0 & 0.0 & 0.0 & 0.0 & 0.0   \\
Kendall &  -.356 & -.339 & -.296 & -.320 & -.298 & -.297\\
p-value & 0.0 & 0.0 & 0.0 & 0.0 & 0.0 & 0.0 \\
  \hline
 \end{tabular} \caption{Examples of high negative Spearman and Kendall Tau rank correlation correlation that are statistically significant. TH, RH, LE, RK, MH, and RW denote top of head, right-hip, left-elbow, right-knee, mid-head, and right wrist respectively, and (C,A) denotes the camera and action. }
 \label{Table:Stats5}
\end{table}

From Table \ref{Table:Stats3}, \ref{Table:Stats4}, and \ref{Table:Stats5} we see a wide variation of correlations and significance levels. The whole dataset is partitioned by the 15 actions, 4 cameras, and 17 joints whose correlation statistics and significance levels were calculated. We demonstrate above the extremes that are witnessed within the model that can be attributed to the inherent noise to depth map estimation, which is not robust in the presence of lighting and occlusion -- to name a few. Nonetheless, a large portion of the sub-samples have moderately high correlation that are statistically significant at all levels. We note that around 80 percent of the joints when partitioned by action and camera have statistically significant correlation at all levels, while 10 percent is significant at $\alpha = 0.05$ and the remaining 10 is either insignificant or significant at a higher level. 

We note that it is of interest that a small portion of the correlation values, $\sim 10 \% $, were negative. This is not intuitive given that depth should increase relative to the $z$-coordinate increasing. We do note that most depth estimation algorithms are susceptible to noise under varying lighting conditions, i.e. lighting is non-uniform. Thus, these negative correlations can attest to the need for improvement in depth map prediction. Furthermore, another plausible and likely scenario is that in the presence of occlusions an occluded joints depth would have a smaller predicted value. 

The overall conclusion of our statistical correlation analysis is that depth values for most actions and cameras have at least weak correlation with a large portion having moderate correlation as defined by correlation $> 0.3$. We hypothesize that even the wide range of correlations is a substantial contributor to the MJPJE reduction seen by our model, relative to the previous methods. In future work, we aim to develop a probabilistic model to assess a chance of occlusion (perhaps by analyzing skeleton angles and camera positions). High probability values of occlusion will enable us to remove corresponding data points to improve the assessment of the correlation. Furthermore, we believe a concentrated effort by the community to generate depth maps that highly correlate with the camera coordinates $z$-value would lead to a substantial reduction in MJPJE.

\begin{figure}[H]
    \caption{Correlation vs average mm error example plots, with best fit line demonstrating negative correlation as desired.}
    \begin{center}
    \includegraphics[width=.75 \textwidth]{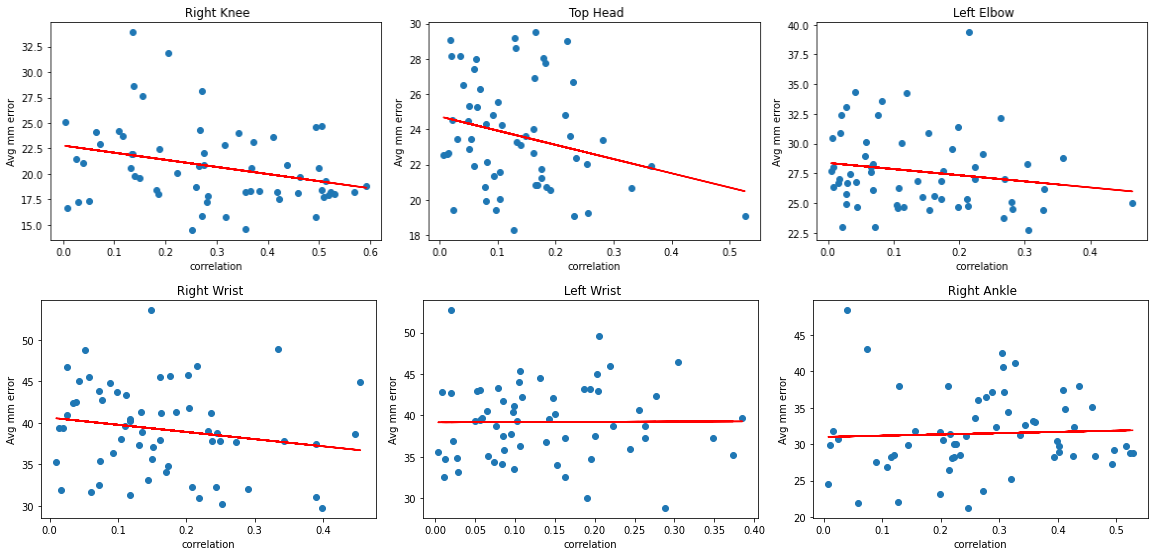}
    \end{center}
\end{figure}

We close the section by commenting with respect to \textbf{Hypothesis 2}. Figure 5 demonstrates negative trend lines between the avg mm and the correlation, i.e. as correlation increase the avg mm error is decreasing. This is indicative that our hypothesis 2 has merit and we believe that further improvement upon depth map estimation, having a dataset with uniform lighting, and omitting occlusions would further support our hypothesis.

\section{Discussion}
This work has begun to uncover the role of depth in the 3d human pose estimation problem. Yet, from both an analytic and statistical perspective more work is needed to fully understand the limit with which depth can play to improve reconstruction. While we study the amount of correlation present between the depth value and the joints $z$-coordinate, further investigation is required to understand what is occurring in individual frames to cause certain joints to have lower correlation as compared to others. We hypothesize that a large contributor to an individual joint having vastly different correlation values is a result of joint occlusions occurring due to camera location in relation to the action. It is well known that depth maps are imprecise with respect to occlusions. The presence of additional inputs, e.g. temporal information, can further shrink the search space of the network leading to more robust 3d pose estimation.

We uncovered that in the event  of perfect correlation, i.e. taking the depth values to be the ground truth $z$-coordinates, that our network can achieve 11 mm average joint reconstruction error on the validation set. This value acts as an absolute minimum for reconstruction error. Thus there is room for up to a 70\% improvement for our network. We are optimistic that a better understanding of the depth value/ $z$-coordinate relationship will help close this gap.

\section{Conclusions}
We propose a novel framework for single person 3d human pose estimation from a single RGB image, which could be considered three-staged. Our framework is flexible and easy to use as the reconstruction network can be paired with any 2d pose estimator and depth map estimator from a monocular image. We provide the following diagram as a visualization of how this pipeline can be implemented into a real time system.

\begin{figure}[H]
  \caption{Diagram outlining how to achieve with our pipeline 3d reconstruction from a monocular image.}
  \begin{center}
  \includegraphics[width=.75 \textwidth]{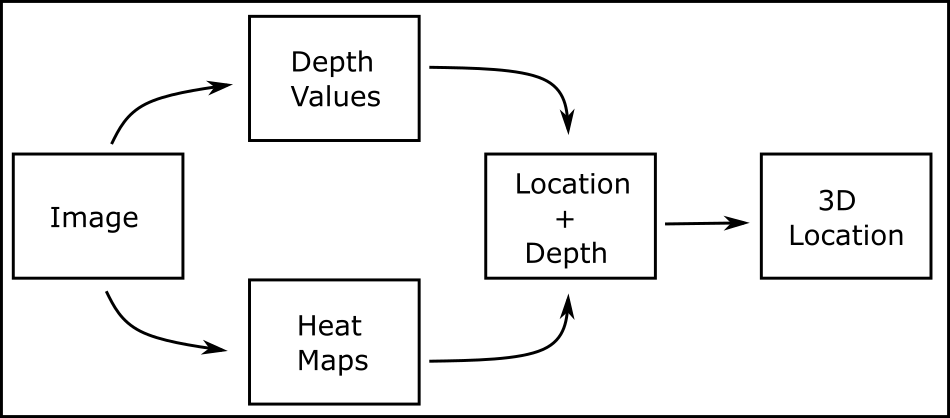}
  \end{center}
\end{figure}

To the best of our knowledge, it is atypical to report the average mm on a per joint basis, so it is difficult to compare our network in these terms to our predecessors. However, given of the substantial reduction we have seen it is relatively safe to assume that on a per joint basis there was a substantial lowering of the per joint errors. We note that joints with the highest correlation values across the entire dataset of $0.345$, e.g. the right-ankle (RA) does not necessarily imply that this joint will have the lowest mm error (RA has error 48.79 under protocol 1 while the minimum is observed at 9.99 for the right and left hips) which had correlation of $.276$. This is easily explainable by the variances of the joint positions which are naturally much larger for extremities compared to joints located near the root.  

The proposed system outperforms previous 3d pose estimators by a significant margin. The height of this error reduction is seen for Protocol 2. We hope that this study invigorates researchers to improve upon the state-of-the-art for depth map estimation so the full potential human pose estimation can be realized. \\

\textbf{Acknowledgement}
This work is part of a University of Iowa Research Foundation invention disclosure \# 21087.0110U1.

\nocite{*}
\bibliographystyle{acm}
\bibliography{Ref}

\end{document}